\documentclass[letterpaper, 10 pt, conference,table]{ieeeconf}  
\usepackage{amsmath} 
\usepackage{bm}
\usepackage{graphicx, float,  subfigure}
\usepackage[linesnumbered,ruled]{algorithm2e}
\usepackage{soul,color}
\usepackage[colorlinks]{hyperref}
\usepackage{xcolor}
\usepackage{cleveref}
\usepackage{mwe}

\newcommand{\bmat}{\begin{bmatrix}}
\newcommand{\emat}{\end{bmatrix}}

\IEEEoverridecommandlockouts                              

\begin{document}



\title{\LARGE \bf
Multi-Profile Quadratic Programming (MPQP) for Optimal Gap Selection and Speed Planning of Autonomous Driving
}

\author{Alexandre~Miranda~A$\tilde{\text{n}}$on$^{1}$,
        Sangjae~Bae$^{1}$,
        Manish~Saroya$^{1}$,
        David~Isele$^{1}$%
\thanks{$^{1}$Honda Research Institute, USA, Inc. San Jose, CA 95134, USA
        {\tt\small \{alexandre\_mirandaanon, sbae, manish\_saroya, disele\}@honda-ri.com}}%
}

\maketitle

\begin{abstract}
Smooth and safe speed planning is imperative for the successful deployment of autonomous vehicles. This paper presents a mathematical formulation for the optimal speed planning of autonomous driving, which has been validated in high-fidelity simulations and real-road demonstrations with practical constraints. The algorithm explores the inter-traffic gaps in the time and space domain using a breadth-first search. For each gap, quadratic programming finds an optimal speed profile, synchronizing the time and space pair along with dynamic obstacles. Qualitative and quantitative analysis in Carla is reported to discuss the smoothness and robustness of the proposed algorithm. Finally, we present a road demonstration result for urban city driving.

\end{abstract}


\section{Introduction}

Trajectory planning has been a vibrant area of research 
engagement in autonomous driving. While foundational planning algorithms have existed for decades \cite{dubins1957curves,reeds1990optimal}, calls to navigate efficiently, precisely, and smoothly through highly dynamic and uncertain scenes where safety is critical have pushed researchers to probe the limits of existing algorithms in an effort to prioritize often conflicting objectives \cite{heilmeier2019minimum,bouton2020reinforcement,bae2022lane}. 
Due to both their interpretability and theoretical guarantees, formal optimization methods such as model predictive control (MPC) and quadratic programming (QP) still remain popular choices
for trajectory planning and control tasks for autonomous
driving \cite{ivanovic2020mats,tariq2023risk}. 

Motion planning in the presence of dynamic and uncertain agents can result in highly non-convex problems \cite{fraichard1993path, gupta2023interaction}. These can either 1) be slow to solve thereby limiting the optimality or time horizon of the solution, or 2) make use of approximations that get stuck in local optima, resulting in shaky or otherwise undesirable trajectories \cite{mohseni2019interaction}. Thus, path and speed decomposition methods have been actively adopted by relieving planning subtasks in terms of complexity \cite{kant1986toward, fraichard1993path, pham2014planning, liu2017speed}. By decoupling the path planning problem from the speed planning problem, we can restrict the optimization space, resulting in faster high-quality solutions \cite{yang2021path}.

\begin{figure}
    \centering
    \includegraphics[width=\columnwidth]{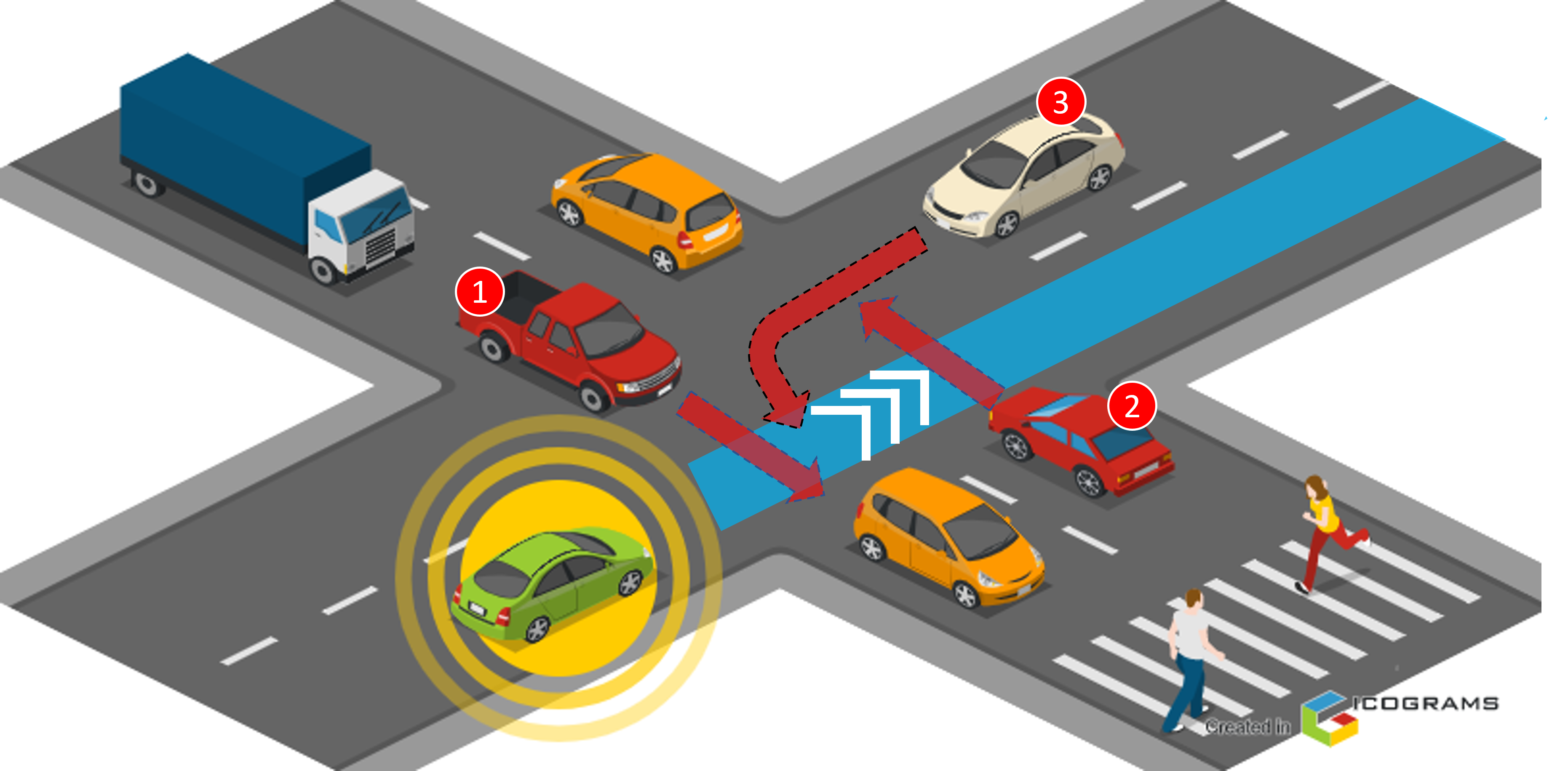}
    \caption{Motivational example: there exists multiple timing options of crossing the intersection for the ego vehicle (in green) along with the crossing timings of cars 1, 2, and 3. Precisely synchronizing the timing and distance as well as choosing the best gaps is the key for a safe maneuver.}
    \label{fig:mpqp_motivation}
    \vspace{-3mm}
\end{figure}

For decades, there have been collective efforts to improve the decomposition method \cite{pham2014planning, jain2019collision}. Especially for autonomous driving applications, a space-time (ST) graph \cite{Xu2022motion, kruger2019graph} has been a popular choice to refine lower and upper bounds, thus reducing the search space. The ST graph also helps address dynamic obstacles and their associated ordering problems \cite{mavrogiannis2020implicit} (see the motivational example in Fig.~\ref{fig:mpqp_motivation}). 

While the ST graph simplifies the problem, the synchronization of time and space remains challenging~\cite{micheli2023nmpc}. To address that, the authors in \cite{liu2017speed} added synchronized time as a decision variable and solved the non-convex problem with iterative QP approximations. However, the iterative process is often computationally heavy and approximations may lead to local optima \cite{boyd2004convex}. Similarly, the author in \cite{Xu2022motion} employed an ST graph and approximated the non-convex problem as QP, but solved it in the space domain to ease the integration of position-based speed limits (e.g., for curved paths). Yet, planning in the space domain cannot seamlessly consider zero speed as it causes the time evaluation to be infinity \cite{sun2020optimal}, problematic in stop-and-go scenarios. In  \cite{shi2023trajectory, kruger2019graph} the authors applied Dynamic Programming (DP) to directly optimize non-convex problems on the ST graph. However, DP typically suffers from the curse of dimensionality, and thus is often limited in number of states and control inputs. Customization \cite{shi2023trajectory} and heuristic solutions \cite{kruger2019graph} (e.g., using spiral curve interpolations \cite{shi2023trajectory}) are inevitable for real-time implementation, while the number of states and controls remain limited. Overall, the existing methods are promising and mathematically sound, but have yet to address rigorous ways of designing speed planners that simultaneously: (i)~search solutions in temporal space, (ii)~secure global optimality without approximations, (iii)~have high computational efficiency without heuristics, and (iv)~address space-time synchronization along with dynamic obstacles. Importantly, thorough validations in various scenarios, often missing in academic literature, are essential before deploying a method in an automated vehicle. 

Thus, we further explore a rigorous speed planner design by encompassing collective efforts in optimal speed planning under the path-speed decomposition scheme. We adopt the space-time graph to refine the lower and upper bounds, while the refinement is along with the temporal steps. This allows for a direct and efficient integration of lower and upper bounds into a QP. The QP, as a convex optimization, yields a globally optimal solution that secures comfort and safety. While robust, hard constraints often lead to overly conservative behaviors (if not infeasible solutions) that leave a real-time system with no course of action. We choose to combine hard and soft constraints to produce behaviors that remain safe while dynamically adapting to the complexities of the current scene. 

In short, this paper contributes to the literature by presenting a mathematical framework that (i) explores multiple combinations of spatial and temporal scenarios associated with dynamic obstacles and (ii) optimizes the speed profile for the scenarios explored. Importantly, the formulation remains in the temporal domain and addresses the space-time synchronization in a convex optimization. We also introduce techniques to integrate lateral acceleration limits along with curvatures on a given path (and thus account for kinematic limits). The proposed framework has been heavily validated in both simulation and road tests. 

\begin{figure*}
    \centering
    \includegraphics[width=\textwidth]{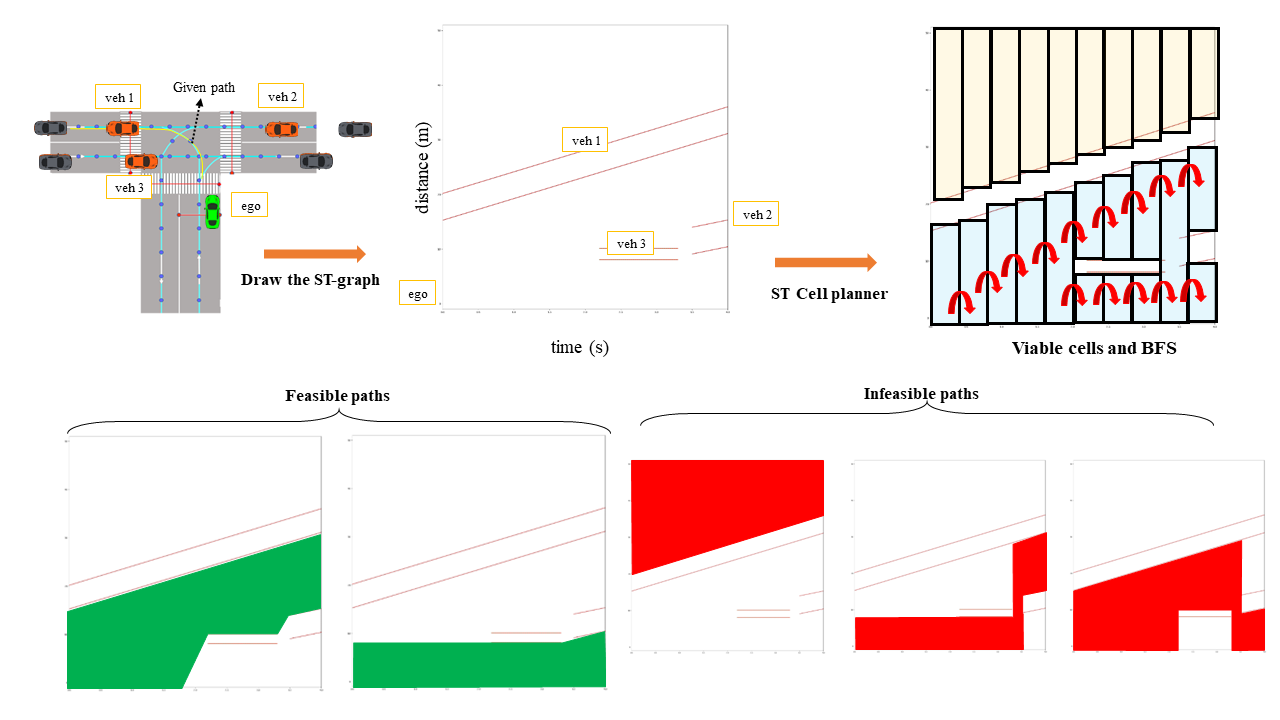}
    \vspace{-3mm}
    \caption{T-junction scenario (top left), corresponding ST-graph (top center) and ST Cell Planner (top right). ST Cell Planner shows also the possible paths found using the Breadth First Search Algorithm (BFS). Bottom images shows the feasible and infeasible paths found in the ST-graph.}
    \label{fig:st-graph-scenario}
    \vspace{-3mm}
\end{figure*}

\section{Space-time (ST) Graph Construction and Domain Initialization}\label{sec:st_graph}
\subsection{Overview}
This section details the algorithm that sequentially (i) generates the space-time graph, (ii) finds cases of passage orders among traffic agents with respect to the ego vehicle (we call it ``\textbf{profile}''), and (iii) obtains the lower and upper bounds associated with each profile. We assume the knowledge of the current state, path-to-follow, and predictions on other traffic.

To illustrate the algorithm, see Fig.~\ref{fig:st-graph-scenario}. Given the ego car's path (in yellow), the predictions on other traffic (with constant speed predictions) indicate the ego vehicle intersecting with three vehicles (veh 1, veh 2, and veh 3). The vehicles are shaped as a polygon and we draw a (time $t$, distance $s$) trajectory where the polygon intersects the given path -- thus, the ST graph. As an instance of the ST graph, the plot at the top center conveys: veh 1 currently intersects with the path and will drive along the path; veh 2 is approaching with a lower speed and will merge into the path later time; veh 3 is closest to the ego vehicle and will cross the path in a very low speed. The ST Cell Planner then segmentizes the continuous ST graph and finds ``viable cells'' (representing a discretized feasible space). Within the segmentation, Breadth-First Search (BFS) finds the possible cases of passage order with respect to the ego vehicle (i.e., profile). Given the profile, the associated lower and upper bounds are generated considering the kinematic limits (green space in the bottom left plot).


When multiple profiles exist, the best profile is chosen with respect to the minimum cost of QP. Note that the profiles are independent of each other, thus applicable for parallel computations.


\subsection{Construction of Space-Time Graph}\label{sec-st-graph}
\label{sec:construct_st_graph}

Aligned with the prior efforts in \cite{shi2023trajectory,Xu2022motion}, the construction of the ST graph is an essential step for MPQP as it is the base for determining the feasible space of the quadratic programming in a later step. 

For each agent, we compute a polygon based on its dimensions and predicted positions. This polygon can be oversized by parameterized margins or uncertainty measures (e.g., noise, risk, or prediction noise). If the polygon intersects with the path of the ego vehicle, the intersecting area (time and distance) is marked on the ST graph. Note that the ST graph is in Frenet frame \cite{werling2010optimal}, and thus we convert the intersecting area to Frenet frame\footnote{Note that we use ``distance'' and ``space'' interchangeably in Frenet frame.}. After all, each agent is represented as two line segments in the (time, distance) coordinate. Note that these lines can be shifted to account for safety margins. 







\subsection{Viable Cell Generation}\label{sec-st-cell-planner}

A ``viable cell'' represents a feasible set in the ST graph, excluding the space occupied by other agents (represented by ``occupied cells''). At each time segment, a cell is defined by the bounds in distance $s$, i.e., $\text{cell} = \{s_{\min}, s_{\max}\}$. For brevity, we call the algorithm of the segmentation and cell generation a ``Cell Planner''. The cell planner first identifies occupied cells associated with each agent at each time segment. It is possible that multiple occupied cells exist. If the occupied cells overlap, they are combined. The viable cells are then found as a complementary set of the occupied cells. For instance, at the time segment with $t=2$, suppose there exist three vehicles (i.e., three occupied cells):
\begin{equation}
  \text{occupied cells} (t = 2) =
    \begin{cases}
      oc_1 = \{4, 6\} \\
      oc_2 = \{5, 8\} \\
      oc_3 = \{20, 25\}.
    \end{cases}       
\end{equation}
As $oc_1$ and $oc_2$ are overlapped, they are combined:
\begin{equation}
  \text{occupied cells} (t = 2) =
    \begin{cases}
      \widetilde{oc_1} = \{4, 8\} \\
      \widetilde{oc_2} = \{20, 25\}
    \end{cases}
\end{equation}
The viable cells are the complementary set of $\widetilde{oc_1}$ and $\widetilde{oc_2}$:
\begin{equation}
  \text{viable cells} (t = 2) =
    \begin{cases}
      vc_1 = \{0, 4\} \\
      vc_2 = \{8, 20\} \\
      vc_3 = \{25, \Bar{s}\}
    \end{cases}       
\end{equation}
where the $\Bar{s}$ is the upper bound in distance.




\subsection{Profile Search}\label{sec:bfs}
We adopt BFS \cite{bundy1984breadth} to discover possible profiles in the ST graph. The search process reads: 
\begin{enumerate}
    \item Starting from the origin, progress one step forward in time.
    \item Expand a profile for each viable cell if the cell overlaps with any viable cells in the previous step. \label{bfs:step2}
    \item Otherwise, terminate and discard the profile.
    \item If the time step reaches the end (i.e., planning horizon), return all valid profiles. 
\end{enumerate}
The step \ref{bfs:step2} indicates that multiple viable cells can be connected in each time step (see the top right plot in Fig.~\ref{fig:st-graph-scenario}). If no viable cell exists at the origin, the ST graph is not traversable, and thus no profile exists.

\subsection{Integration to QP}\label{sec:st-qp-integration}
Recall, constructing the ST graph is eventually to facilitate the underlying speed optimization. There are three techniques we employ: \textbf{First}, we segmentize the ST graph with the equivalent time step ($0.1$ sec) to that of QP. This is straightforward while being highly effective in synchronizing the path and the speed. \textbf{Second}, we generate a lower bound of each profile by connecting the lower end of viable cells at each time step. Similarly, an upper bound is generated by connecting the upper end of viable cells. \textbf{Third}, we post-filter all valid profiles with kinematic limits (in speed, acceleration, and jerk). This helps reduce the number of profiles as well as ensure the (longitudinal) kinematic limits. 

Depending on the step size, the overall process might be computationally expensive. However, processing each profile in Section~\ref{sec:bfs} and \ref{sec:st-qp-integration} is completely separable, thus parallel computations resolve the issue. A resulting profile is represented by a pair of boundaries with synchronized time. That can be seamlessly integrated into QP. 

\section{Optimization problem}\label{sec:optimization}
Given the piecewise linear bounds obtained from the ST graph, we formulate quadratic programming \cite{boyd2004convex} that optimizes speed and keeps safety. 

\subsubsection{System Dynamics}
The state includes [distance, speed, acceleration] denoted by [$p,v,a$], and the control is jerk $j$. We aim to find the optimal speed profile along with the longitudinal direction, and thus we leverage the simplified longitudinal dynamics:
\begin{align}
    p(t+1) &= p(t) + v(t)dt,\quad\forall t\in\left[0,\ldots,N-1\right],\label{eq:p1}\\
    v(t+1) &= v(t) + a(t)dt,\quad\forall t\in\left[0,\ldots,N-2\right],\label{eq:v1}\\
    a(t+1) &= a(t) + j(t)dt,\quad\forall t\in\left[0,\ldots,N-3\right].\label{eq:a1}
\end{align}
With the augmented variable $x=[p,v,a,j]$, we write the dynamics in the compact form:
\begin{equation}
    x(t+1) = f(x(t)),
\end{equation}
where $f(\cdot)$ is the linear mapping with Eqn.~\eqref{eq:p1}-\eqref{eq:a1}.

\subsection{Objective Function}
The objective is to enhance passenger comfort and reduce travel time. Formally, the objective function $F$ reads:
\begin{equation}
    F = \frac{1}{2}x^\top H x + x^\top h,
\end{equation}
where $H \in R^{\dim(x)\times \dim(x)}$ is a diagonal matrix where the weight $w_a$ is imposed on accelerations (for all $t\in\left[0,\ldots,N-2\right]$) and the weight $w_j$ is imposed on jerks (for all $t\in\left[0,\ldots,N-3\right]$) while it is zero everywhere else. The vector $h$ has zero value for all elements except for the final displacement at $t=N$ (i.e., $p(N)$), which is set to $-w_f$ ($w_f>0$) to encourage a large displacement with respect to the initial position (i.e., a shorter travel time).

\subsection{Constraints}
There are two types of constraints: (i) initial conditions and (ii) lower and upper bounds. The initial conditions read:
\begin{align}
    p(0)&=p_0,\\
    v(0)&=v_0
\end{align}
where $p_0$ and $v_0$ denote the displacement and speed measured at the current time, respectively. In compact form, we can rewrite it with the augmented variable $x$, i.e., $x(0)=x_0$.

Recall that the lower and upper bounds are obtained from the ST graph in Section~\ref{sec-st-graph} and thus they are given and known. The corresponding inequality constraints read:
\begin{equation}
    lb(t) \leq x(t) \leq ub(t), \quad\forall t\in\left[0,\ldots,N\right]. \label{eq:bounds}
\end{equation}
Note that the lower and upper bounds can be engineered with customized margin choices (such as parameters), which can potentially cause infeasibility issues. Details are discussed in Section~\ref{sec:soft-constraint}.

\subsection{Complete problem}
The complete optimization problem reads:
\begin{subequations}
\begin{align}
    \min_x &\frac{1}{2}x^\top H x + x^\top h\\
    &\text{subject to:}\nonumber\\
    x(t+1)&=f(x(t)), \quad\forall t\in\left[0,\ldots,N-1\right],\\
    lb(t) &\leq x(t) \leq ub(t), \quad\forall t\in\left[0,\ldots,N\right],\\
    x(0)&=x_0.
\end{align}
\end{subequations}
Note that this is a generic form of QP without any (iterative) approximations -- it retains nice convex optimization properties \cite{boyd2004convex}, e.g., optimality and convergence rate.

\begin{figure}[t]
    \centering
    \includegraphics[width=1\columnwidth]{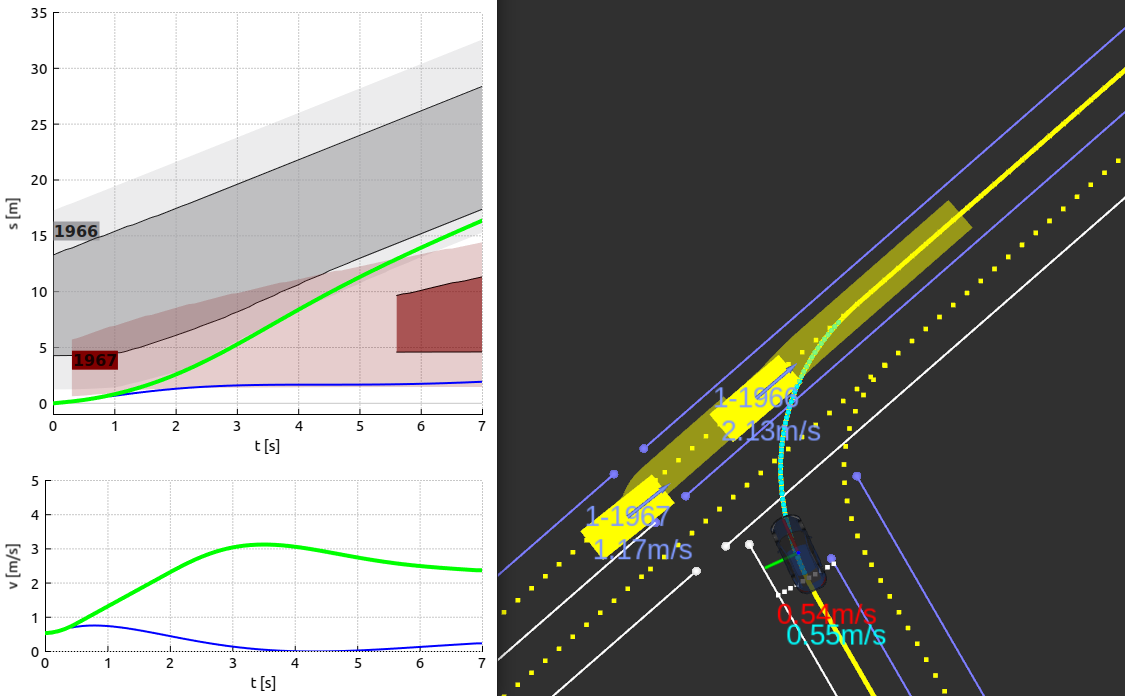}
    \caption{T junction scenario with 2 agents (Car 1966 in grey, Car 1967 in dark red). The hard constraints can be seen in a solid color fill, the soft constraints can be seen in a semi-transparent color fill.} 
    \label{fig:soft_const_example}
    \vspace{-3mm}
\end{figure}

\subsection{Soft-constraints for relaxed problem}\label{sec:soft-constraint}
The inequality constraints \eqref{eq:bounds} can be violated depending on the constructed ST graph, leading to a problem that is infeasible to solve. Examples include: (i) two predicted agent trajectories overlapping, causing no-existing feasible throughput within the planning horizon, and (ii) the initial position of the ego vehicle being outside the feasible space defined by the two bounds (due to excessive margin parameters and/or perception/localization noises). Infeasible solutions may cause jerky and uncomfortable maneuvering (depending on the pre-defined actions for infeasible solutions). 

One well-received solution is to reformulate the problem by relaxing the constraints, leaving the problem soft-constrained. We add a slack variable for each bound (i.e., $s_{lb}$ and $s_{ub}$) and penalize non-zero slack variables in the objective function. The problem now reads:
\begin{subequations}    
\begin{align}
    \min_{x,s_{lb},s_{ub}} &\frac{1}{2}x^\top H x + x^\top h + w_{b}^\top(s_{lb}+s_{ub})\\
    &\text{subject to:}\nonumber\\
    x(t+1)&=f(x(t)), \quad\forall t\in\left[0,\ldots,N-1\right],\\
    lb(t)-s_{lb}(t) &\leq x(t) \leq ub(t)+s_{ub}(t), \quad\forall t\in\left[0,\ldots,N\right],\\
    x(0)&=x_0,\\
    0&\leq s_{lb}(t) \leq s_{\max lb}(t),\\
    0&\leq s_{ub}(t) \leq s_{\max ub}(t),
\end{align} \label{eqn:qp_soft}
\end{subequations}
where $w_b$ denotes the penalty weight on the boundary slack variables. Note that the slack variables are temporal for all $t$ and only non-zero values will be penalized. Also, note that the slack variables are bounded (by $s_{\max lb}$ and $s_{\max ub}$) to ensure safety.
Figure~\ref{fig:soft_const_example} illustrates the soft margins for a simple T-Junction scenario.

\subsection{Funnel Technique to Impose Lateral Acceleration Limits}
One caveat of longitudinal speed planning is that it disregards lateral accelerations and associated kinematic limits associated with the given path. However, directly evaluating the lateral accelerations leads the problem to non-convex, resulting in a higher complexity problem to solve \cite{Xu2022motion, liu2017speed}. We introduce an easily implementable yet effective method: the funnel technique. 

Given the path, we compute the curvature $K$ at each position step $s$. With a parametric lateral acceleration limit $a_\ell$, the associated speed limit is: 
\begin{equation}
    v_s = \sqrt{\frac{a_\ell}{K_s}}.
\end{equation}
With a parametric distance lookahead $L$, the speed limit $v_\ell$ is set to the minimum of $v_s$, i.e., $v_\ell=\min(v_s,v_\text{max}), \forall s\in\{0,\ldots,L\}$, where $v_\text{max}$ is the traffic speed limit. Yet, the speed limit $v_\ell$ cannot be immediately used to upper bound the speed in QP in Eqn.~\eqref{eqn:qp_soft} since the QP becomes infeasible if the current speed is higher than $v_\ell$. Instead, the upper bound is funneled down from the traffic speed limit to $v_\ell$ by linearly decreasing with a fixed rate. Note that a high decreasing rate might overly tighten the upper bounds (which in turn restricts the feasible set). We found that half the minimum acceleration, i.e., $-\frac{a_\text{min}}{2}$, is a comfortable rate.


\begin{figure*}
\centering
\includegraphics[width=.5\columnwidth]{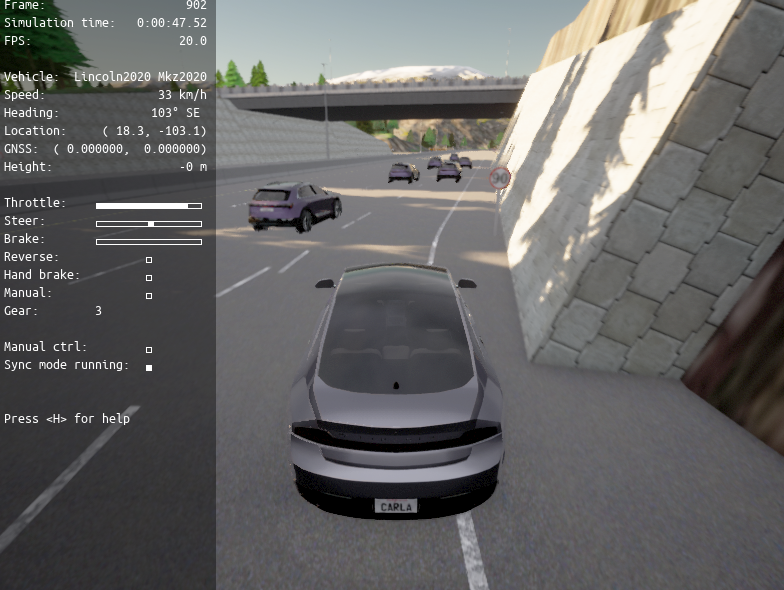}
\includegraphics[width=.5\columnwidth]{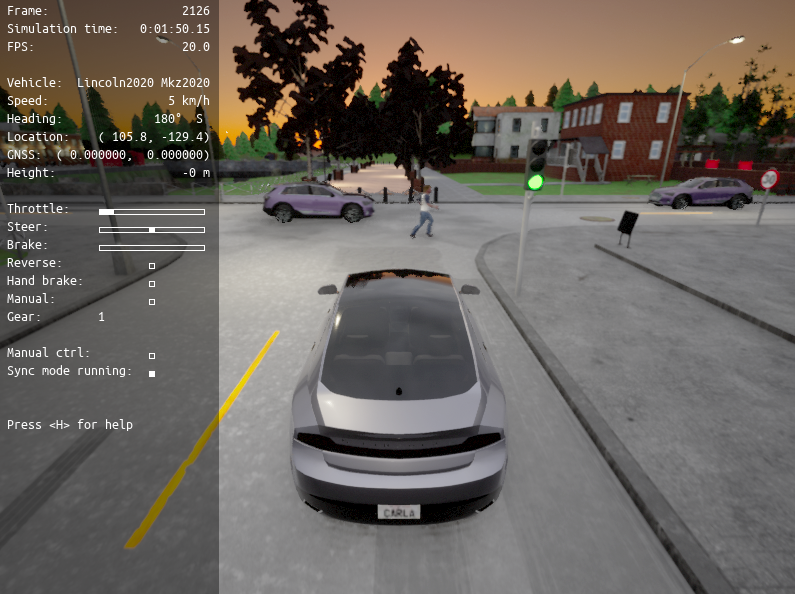}
\includegraphics[width=.5\columnwidth]{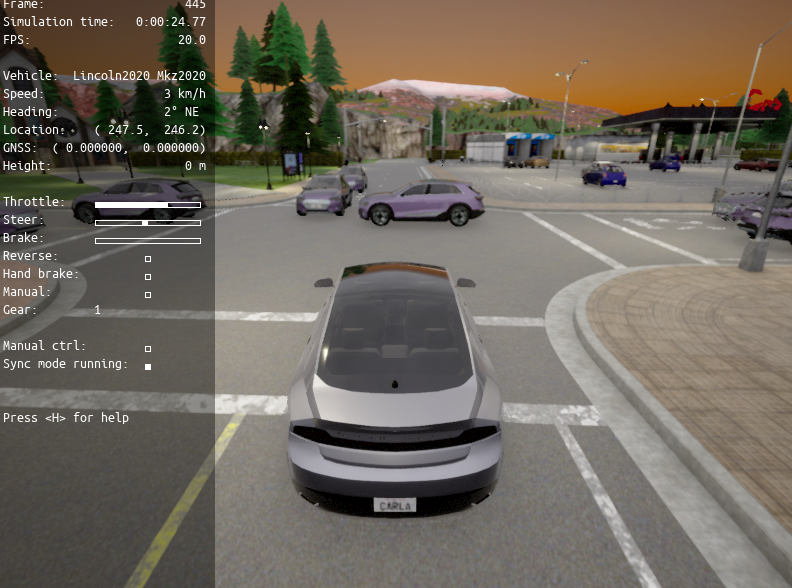}
\includegraphics[width=.5\columnwidth]{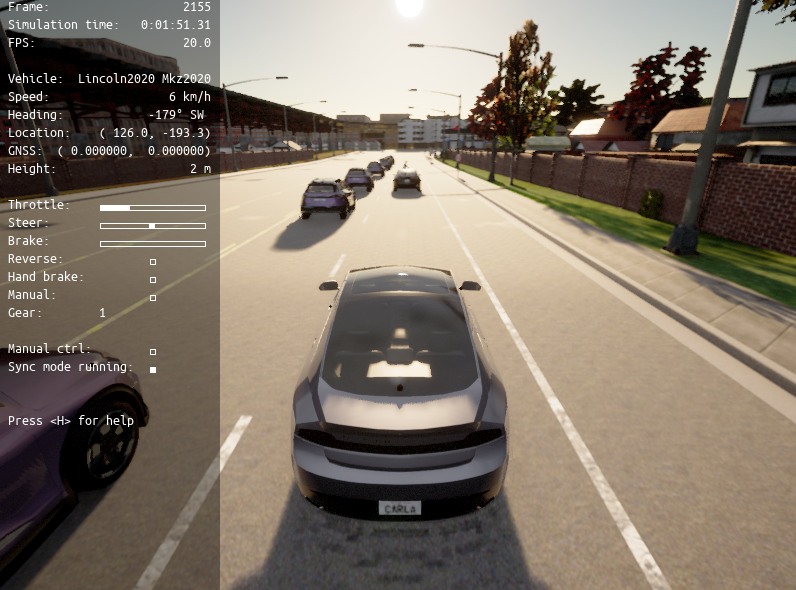}
\caption{Testing scenarios (from left): (i) highway merging, (ii) T-junction, (iii) four-way intersection, and (iv) lane changing in dense traffic.}
\label{fig:scenarios}
\end{figure*} 

\begin{table*}[t]
\centering
\begin{tabular}{clllllllllllll}
\hline
\textbf{Scenario} & \multicolumn{1}{c}{\textbf{Time}} & \multicolumn{1}{c}{\textbf{\begin{tabular}[c]{@{}c@{}}Succ.\\ (\%)\end{tabular}}} & \multicolumn{1}{c}{\textbf{\begin{tabular}[c]{@{}c@{}}Coll.\\ (\%)\end{tabular}}} & \multicolumn{1}{c}{\textbf{\begin{tabular}[c]{@{}c@{}}Time\\ out\\ (\%)\end{tabular}}} & \multicolumn{1}{c}{\textbf{\begin{tabular}[c]{@{}c@{}}Brake\\ Avg\end{tabular}}} & \multicolumn{1}{c}{\textbf{\begin{tabular}[c]{@{}c@{}}Thr.\\ Avg\end{tabular}}} & \multicolumn{1}{c}{\textbf{\begin{tabular}[c]{@{}c@{}}Acc.\\ Max\end{tabular}}} & \multicolumn{1}{c}{\textbf{\begin{tabular}[c]{@{}c@{}}Brake\\ Jerk\\ Avg\end{tabular}}} & \multicolumn{1}{c}{\textbf{\begin{tabular}[c]{@{}c@{}}Thr.\\ Jerk\\ Avg\end{tabular}}} & \multicolumn{1}{c}{\textbf{\begin{tabular}[c]{@{}c@{}}Ang.\\ Acc.\\ Avg\end{tabular}}} & \multicolumn{1}{c}{\textbf{\begin{tabular}[c]{@{}c@{}}Ang.\\ Acc.\\ Max\end{tabular}}} & \multicolumn{1}{c}{\textbf{\begin{tabular}[c]{@{}c@{}}Ang.\\ Jerk\\ Avg\end{tabular}}} & \multicolumn{1}{c}{\textbf{\begin{tabular}[c]{@{}c@{}}Ang.\\ Jerk\\ Max\end{tabular}}} \\ \hline
\rowcolor[HTML]{EFEFEF} 
Highway merging & 20.17 & 100 & 0 & 0 & -0.64 & 0.75 & 1.87 & -0.73 & 0.9 & 1.36 & 8.22 & 2.31 & 15.43 \\
\rowcolor[HTML]{FFFFFF} 
T-junction & 22.18 & 100 & 0 & 0 & -0.71 & 0.66 & 1.97 & -0.86 & 0.85 & 2.44 & 12.26 & 3.46 & 23.45 \\
\rowcolor[HTML]{EFEFEF} 
\begin{tabular}[c]{@{}c@{}}Four-way \\ intersection\end{tabular} & 31.34 & 100 & 0 & 0 & -0.32 & 0.42 & 1.96 & -0.47 & 0.45 & 0.97 & 10.59 & 4.24 & 47.36 \\
\rowcolor[HTML]{FFFFFF} 
\begin{tabular}[c]{@{}c@{}}Lane change\\ in dense traffic\end{tabular} & 41.77 & 100 & 0 & 0 & -0.19 & 0.25 & 1.4 & -0.34 & 0.32 & 1.5 & 18.84 & 2.17 & 30.29 \\ \hline
\end{tabular}
\caption{Simulation results with the four different scenarios. The result of each scenario is with 50 runs. The identical system is used for all four scenarios, and positions and driving behaviors (e.g., target speed) of the other vehicles are randomly initialized in each run. A run stops if a collision occurs or a time limit (80 seconds) is reached. }\label{table:result}
\vspace{-7mm}
\end{table*}

\section{Validations}\label{sec:result}
\subsection{Simulation Overview}
We leverage CARLA \cite{Dosovitskiy17} for the high-fidelity simulation environment which has been broadly adopted for validating autonomous driving research. Scenarios are generated using the ``scenario runner'', the CARLA built-in simulation generation platform. 

We test MPQP with four different scenarios: (i) highway merging, (ii) T-junction, (iii) four-way intersection, and (iv) lane changing in dense traffic, as presented in Fig.~\ref{fig:scenarios}. Each scenario has different road structures, target lanes, and traffic participants. 

\noindent \textbf{Highway merging scenario:} The ego vehicle starts on the ramp and merges into the lane. Other vehicles in the target lane drive fast with randomness in their yielding behaviors toward the ego vehicle. 

\noindent \textbf{T-junction scenario:} The ego vehicle makes an unprotected left turn while maintaining safety for crossing pedestrians. 

\noindent \textbf{Four-way intersection scenario:} The ego vehicle makes a left turn. The increased complexity is due to additional vehicles approaching from the opposite direction. 

\noindent \textbf{Lane changing scenario:} The goal is to lane change into dense traffic to avoid an emergency vehicle stopped in front. There exists no safe space for the ego vehicle to merge without interacting with other cars (i.e., without the cooperation of other vehicles). 

This set of scenarios and their inherent challenges evaluate the general performance of the proposed algorithm. Note that we use the same algorithm for all scenarios, while positions and driving behaviors (e.g., target speed) of other vehicles are randomly initialized in each run. Other vehicles are modeled as intelligent driver model (IDM) \cite{kesting2010enhanced} with some modifications to detect the ego vehicle in the adjacent lane. The yielding behavior is triggered according to a pre-set Bernoulli parameter. Interested readers are referred to \cite{bae2022lane} for details. Overall, the vehicles are designed to be aggressive in order to challenge our algorithm. The behavior model for the agents is lane following (except for pedestrians that randomly change directions) and traffic does not change lanes. 


MPQP's processing time averages 20 ms with a maximum time cycle of 84 ms on a laptop with an Intel® Core™ i9-12900H and 32 GB RAM. This processing time includes all the aforementioned steps: ST graph construction (using ClipperLib \cite{angusjClipper2Polygon}), ST Cell Planning, and quadratic programming (using qpOASES \cite{ferreau2014qpoases}). While the planned trajectory is sent to an underlying controller, MPQP finds a new solution. The planning horizon was set to 10 seconds for slow-speed scenarios and 15 seconds for high-speed scenarios. This indicates a strong potential for MPQP to be applicable to real-time systems even with a relatively long planning horizon. 

\subsection{Simulation Result}
\begin{figure}
\centering
\includegraphics[width=.49\columnwidth]{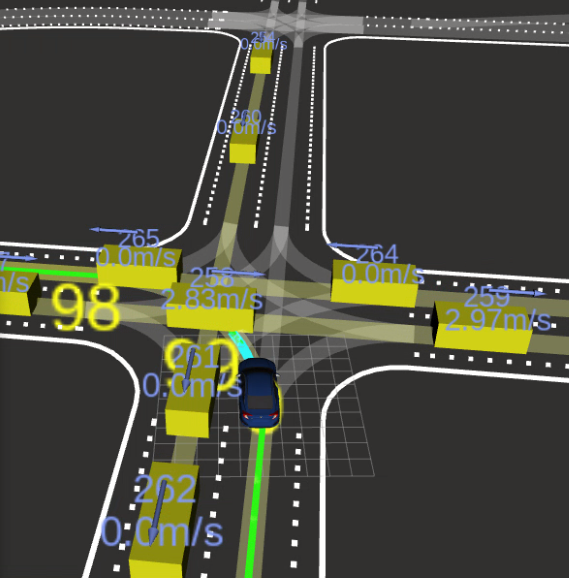}
\includegraphics[width=.49\columnwidth]{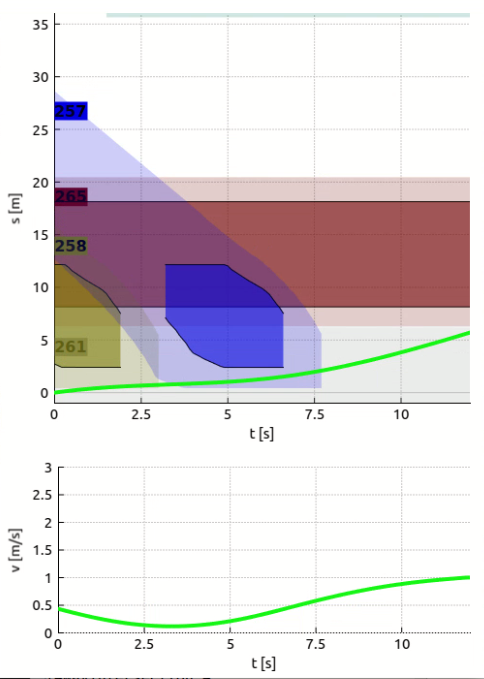}
\caption{Testing scenarios (from left): (i) highway merging, (ii) T-junction, (iii) four-way intersection, and (iv) lane changing in dense traffic.}
\label{fig:4way}
\vspace{-3mm}
\end{figure}

We first investigate the performance of MPQP in the four-way intersection scenario aligned with the motivational example in Fig.~\ref{fig:mpqp_motivation}. Unlike the motivational example, the ego vehicle takes a left turn to interact with unexpected pedestrian crossing. The general behaviors of MPQP were observed as: (i) the ego vehicle approaches the intersection; (ii) the ego vehicle stays stopped until the oncoming vehicles cross - Figure~\ref{fig:4way} illustrates the ego vehicle slowly cruising until the oncoming cars (labeled as 258 and 257) pass on lane 98; (iii) if the car (car 264 in Fig.~\ref{fig:4way}) on the merging lane does not slow down, the ego vehicle slows down and waits. Otherwise, it proceeds. (iv) While completing the turn, a pedestrian suddenly crosses the road and MPQP waits. Depending on the initial positions and aggressiveness of the traffic, the behaviors might vary, however, MPQP consistently secured safe and successful merging. 

Now, we extend our analysis across the aforementioned four scenarios: highway merging, T-junction, four-way intersection, and lane change. Table~\ref{table:result} tabulates the overall performance for each scenario that was repeated 50 times with random initialization. Overall, MPQP successfully completes all scenarios without collisions or timeout. Note that MPQP does not have prior knowledge of what scenario will be running and of what behavior model the traffic is based upon. That is, MPQP demonstrates a capability of generalizability without situational knowledge and with the presence of uncertainties. Some observations from the results: (i) Interestingly, the T-junction scenario has a higher average on the brake than that in the highway (which has a higher speed throughput). This is explained by the uncertain behavioral changes of traffic in the T-junction scenario. Due to the unexpected acceleration of the vehicle at the merging point, the ego vehicle had to slow down strongly to keep the safety. In fact, the same behavior is observed in the four-way scenario, however, the average speed is much lower due to the density which results in less aggressive brake and jerk. 
(ii) The ego vehicle tends to drive smoother and slower for lane changing in dense traffic, waiting for their cooperation (i.e., making room for the ego vehicle). (iii) The ego vehicle tends to be more conservative in the T-junction and four-way scenarios as there exists priority between traffic agents. The priority is given to the pedestrians and oncoming cars (especially in the T-junction scenario). Accordingly, the ego vehicle tends to be defensive and merges to the target lane only if the car obviously gives way to the ego vehicle or if safety is guaranteed (e.g., no vehicle). The overall behaviors are aligned with an experienced human driver who balances between being defensive and not being too conservative.

\subsection{On-road Demonstration}
\begin{figure}[t]
    \centering
    \includegraphics[width=0.8\columnwidth]{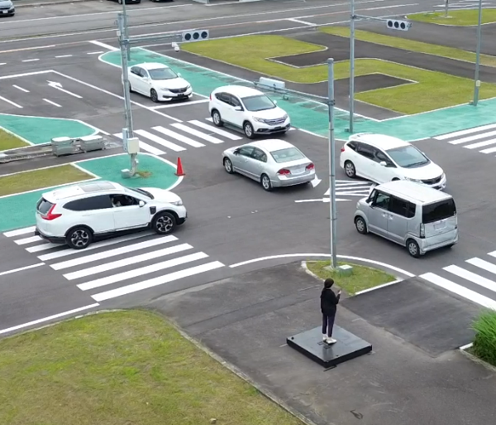}
    \caption{Public demonstration of MPQP on a closed course in Joso city, Japan. The white CR-V on the left of the image is the AD vehicle.}
    \label{fig:joso}
    \vspace{-3mm}
\end{figure}

While the simulation effectively evaluates the general performance of MPQP, there still exists a sim-to-real gap. That is, the behavior model of other vehicles often fails to correctly represent the decision making process of other drivers. Simulation environments do not precisely represent localization and perception noises. Different controllers are used, and there are variations in the physics and system timing. 
We thus extended our validation in the real-world urban city course in Joso city, Japan, 
in July 2023. The course includes several driving scenarios, including 4 way intersection (snapshot in Fig.~\ref{fig:joso}), T-junction, and lane change. 

Public participants feedback was generally positive. Participants mentioned the ride was very smooth and had good user comfort. Some users said the crowded intersection felt complicated even for human drivers. Some participants said some maneuvers, like the lane change and driving around a stopped car, felt a little dangerous but expressed that they had trust in the autonomous driving system. 

\section{Conclusion}\label{sec:conclusion}
Inspired by the efficacy of path-speed decomposition methods, this paper provides a mathematical framework to optimally plan a speed profile for automated vehicles, namely MPQP. Given a path, we leverage the Breadth First Search algorithm to find timing combinations to follow the path when dynamic objects are present. Each profile is then integrated into Quadratic Programming as lower and upper bounds. The formulation and implementation are computationally efficient, being capable of providing a new solution within 20 ms on average for 10+ seconds of planning horizon. Simulation results in CARLA demonstrate the strong potential of MPQP, followed by a real-car demonstration with public participants. One caveat is that this work relies on deterministic predictions for traffic which precludes the ability to yield interactive behaviors. Handling uncertainties in multi-modal predictions remains for future work.

\section*{Acknowledgement}
The authors would like to thank Dr. Yuji Yasui and Dr. Takayasu Kumano at Honda R\&D Co., Ltd., Japan, for their general support for the research and vehicle test. 




\newpage


\bibliographystyle{unsrt}
\bibliography{reference.bib}

\begin{thebibliography}{10}

\bibitem{dubins1957curves}
Lester~E Dubins.
\newblock On curves of minimal length with a constraint on average curvature, and with prescribed initial and terminal positions and tangents.
\newblock {\em American Journal of mathematics}, 79(3):497--516, 1957.

\bibitem{reeds1990optimal}
James Reeds and Lawrence Shepp.
\newblock Optimal paths for a car that goes both forwards and backwards.
\newblock {\em Pacific journal of mathematics}, 145(2):367--393, 1990.

\bibitem{heilmeier2019minimum}
Alexander Heilmeier, Alexander Wischnewski, Leonhard Hermansdorfer, Johannes Betz, Markus Lienkamp, and Boris Lohmann.
\newblock Minimum curvature trajectory planning and control for an autonomous race car.
\newblock {\em Vehicle System Dynamics}, 2019.

\bibitem{bouton2020reinforcement}
Maxime Bouton, Alireza Nakhaei, David Isele, Kikuo Fujimura, and Mykel~J Kochenderfer.
\newblock Reinforcement learning with iterative reasoning for merging in dense traffic.
\newblock In {\em 2020 IEEE 23rd International Conference on Intelligent Transportation Systems (ITSC)}, pages 1--6. IEEE, 2020.

\bibitem{bae2022lane}
Sangjae Bae, David Isele, Alireza Nakhaei, Peng Xu, Alexandre~Miranda A{\~n}on, Chiho Choi, Kikuo Fujimura, and Scott Moura.
\newblock Lane-change in dense traffic with model predictive control and neural networks.
\newblock {\em IEEE Transactions on Control Systems Technology}, 2022.

\bibitem{ivanovic2020mats}
Boris Ivanovic, Amine Elhafsi, Guy Rosman, Adrien Gaidon, and Marco Pavone.
\newblock Mats: An interpretable trajectory forecasting representation for planning and control.
\newblock {\em arXiv preprint arXiv:2009.07517}, 2020.

\bibitem{tariq2023risk}
Faizan~M Tariq, David Isele, John~S. Baras, and Sangjae Bae.
\newblock {RCMS}: Risk-aware crash mitigation system for autonomous vehicles.
\newblock {\em IEEE Intelligent Vehicles Symposium}, 2023.

\bibitem{fraichard1993path}
Thierry Fraichard and Christian Laugier.
\newblock Path-velocity decomposition revisited and applied to dynamic trajectory planning.
\newblock In {\em [1993] Proceedings IEEE International Conference on Robotics and Automation}, pages 40--45. IEEE, 1993.

\bibitem{gupta2023interaction}
Piyush Gupta, David Isele, Donggun Lee, and Sangjae Bae.
\newblock Interaction-aware trajectory planning for autonomous vehicles with analytic integration of neural networks into model predictive control.
\newblock {\em 2023 International Conference on Robotics and Automation (ICRA)}, 2023.

\bibitem{mohseni2019interaction}
Anahita Mohseni-Kabir, David Isele, and Kikuo Fujimura.
\newblock Interaction-aware multi-agent reinforcement learning for mobile agents with individual goals.
\newblock In {\em 2019 International Conference on Robotics and Automation (ICRA)}, pages 3370--3376. IEEE, 2019.

\bibitem{kant1986toward}
Kamal Kant and Steven~W Zucker.
\newblock Toward efficient trajectory planning: The path-velocity decomposition.
\newblock {\em The international journal of robotics research}, 5(3):72--89, 1986.

\bibitem{pham2014planning}
Quang-Cuong Pham, St{\'e}phane Caron, Puttichai Lertkultanon, and Yoshihiko Nakamura.
\newblock Planning truly dynamic motions: Path-velocity decomposition revisited.
\newblock {\em arXiv preprint arXiv:1411.4045}, 2014.

\bibitem{liu2017speed}
Changliu Liu, Wei Zhan, and Masayoshi Tomizuka.
\newblock Speed profile planning in dynamic environments via temporal optimization.
\newblock In {\em 2017 IEEE Intelligent Vehicles Symposium (IV)}, pages 154--159, 2017.

\bibitem{yang2021path}
Zeyu Yang, Jin Huang, Hui Yin, Diange Yang, and Zhihua Zhong.
\newblock Path tracking control for underactuated vehicles with matched-mismatched uncertainties: An uncertainty decomposition based constraint-following approach.
\newblock {\em IEEE Transactions on Intelligent Transportation Systems}, 23(8):12894--12907, 2021.

\bibitem{jain2019collision}
Vasundhara Jain, Uli Kolbe, Gabi Breuel, and Christoph Stiller.
\newblock Collision avoidance for multiple static obstacles using path-velocity decomposition.
\newblock {\em IFAC-PapersOnLine}, 52(8):265--270, 2019.

\bibitem{Xu2022motion}
Wenda Xu.
\newblock {\em Motion Planning for Autonomous Vehicles in Urban Scenarios: A Sequential Optimization Approach}.
\newblock PhD thesis, Carnegie Mellon University, 02 2022.

\bibitem{kruger2019graph}
Till-Julius Kr{\"u}ger, Daniel G{\"o}hring, and Fritz Ulbrich.
\newblock {\em Graph-Based Speed Planning for Autonomous Driving}.
\newblock PhD thesis, Free University of Berlin Berlin, Germany, 2019.

\bibitem{mavrogiannis2020implicit}
Christoforos Mavrogiannis, Jonathan~A DeCastro, and Siddhartha~S Srinivasa.
\newblock Implicit multiagent coordination at unsignalized intersections via multimodal inference enabled by topological braids.
\newblock {\em arXiv preprint arXiv:2004.05205}, 2020.

\bibitem{micheli2023nmpc}
Francesco Micheli, Mattia Bersani, Stefano Arrigoni, Francesco Braghin, and Federico Cheli.
\newblock Nmpc trajectory planner for urban autonomous driving.
\newblock {\em Vehicle system dynamics}, 61(5):1387--1409, 2023.

\bibitem{boyd2004convex}
Stephen~P Boyd and Lieven Vandenberghe.
\newblock {\em Convex optimization}.
\newblock Cambridge university press, 2004.

\bibitem{sun2020optimal}
Chao Sun, Jacopo Guanetti, Francesco Borrelli, and Scott~J Moura.
\newblock Optimal eco-driving control of connected and autonomous vehicles through signalized intersections.
\newblock {\em IEEE Internet of Things Journal}, 7(5):3759--3773, 2020.

\bibitem{shi2023trajectory}
Xiaowei Shi and Xiaopeng Li.
\newblock Trajectory planning for an autonomous vehicle with conflicting moving objects along a fixed path--an exact solution method.
\newblock {\em Transportation Research Part B: Methodological}, 173:228--246, 2023.

\bibitem{werling2010optimal}
Moritz Werling, Julius Ziegler, S{\"o}ren Kammel, and Sebastian Thrun.
\newblock Optimal trajectory generation for dynamic street scenarios in a frenet frame.
\newblock In {\em 2010 IEEE international conference on robotics and automation}, pages 987--993. IEEE, 2010.

\bibitem{bundy1984breadth}
Alan Bundy and Lincoln Wallen.
\newblock Breadth-first search.
\newblock {\em Catalogue of artificial intelligence tools}, pages 13--13, 1984.

\bibitem{Dosovitskiy17}
Alexey Dosovitskiy, German Ros, Felipe Codevilla, Antonio Lopez, and Vladlen Koltun.
\newblock {CARLA}: {An} open urban driving simulator.
\newblock In {\em Proceedings of the 1st Annual Conference on Robot Learning}, pages 1--16, 2017.

\bibitem{kesting2010enhanced}
Arne Kesting, Martin Treiber, and Dirk Helbing.
\newblock Enhanced intelligent driver model to access the impact of driving strategies on traffic capacity.
\newblock {\em Philosophical Transactions of the Royal Society A: Mathematical, Physical and Engineering Sciences}, 368(1928):4585--4605, 2010.

\bibitem{angusjClipper2Polygon}
{C}lipper2 - {P}olygon {C}lipping and {O}ffsetting {L}ibrary --- angusj.com.
\newblock \url{http://www.angusj.com/clipper2/Docs/Overview.htm}.
\newblock [Accessed 27-Jun-2023].

\bibitem{ferreau2014qpoases}
Hans~Joachim Ferreau, Christian Kirches, Andreas Potschka, Hans~Georg Bock, and Moritz Diehl.
\newblock qpoases: A parametric active-set algorithm for quadratic programming.
\newblock {\em Mathematical Programming Computation}, 6:327--363, 2014.

\end{thebibliography}

\end{document}